\documentclass[9pt]{article}
\usepackage{spconf,amsmath,graphicx}
\usepackage{cite}
\usepackage{graphicx} 
\usepackage[linesnumbered,lined,ruled]{algorithm2e}

\usepackage{subfigure}
\usepackage{array}
\usepackage{textcomp}
\usepackage{stfloats}
\usepackage{url}
\usepackage{hyperref} 
\usepackage{helvet}
\usepackage[english]{babel}
\usepackage{color}      
\usepackage{bm}         
\usepackage{multirow}
\usepackage{booktabs}
\usepackage{epstopdf}
\usepackage{epsfig}
\usepackage{pifont}
\usepackage{amsmath,amsthm}
\usepackage{amssymb}
\newtheorem{theorem}{Theorem}
\newtheorem{asmp}{Assumption}
\usepackage{marvosym}   

\title{Disentangle Estimation of Causal Effects from Cross-silo Data}
\name{Yuxuan~Liu\textsuperscript{1},
        Haozhao~Wang\textsuperscript{2},
        Shuang~Wang\textsuperscript{4},
        Zhiming~He\textsuperscript{1},
        Wenchao~Xu\textsuperscript{3},
        Jialiang~Zhu\textsuperscript{1},
        Fan~Yang\textsuperscript{1,5}
        \thanks{*Haozhao Wang is corresponding author}
        }

\address{1:University of Electronic Science and Technology of China, Chengdu, China,\\ 2:Huazhong University of Science and Technology, Wuhan, China,\\ 3:Hong Kong Polytechnic University, Hong Kong, China,4:Nuowei Technology, Hangzhou, China,\\
5: Inner Mongolia Normal University, Hohhot, China}
\begin{document}
%
\maketitle
\begin{abstract}
Estimating causal effects among different events is of great importance to critical fields such as drug development. Nevertheless, the data features associated with events may be distributed across various silos and remain private within respective parties, impeding direct information exchange between them. This, in turn, can result in biased estimations of local causal effects, which rely on the characteristics of only a subset of the covariates.
To tackle this challenge, we introduce an innovative disentangle architecture designed to facilitate the seamless cross-silo transmission of model parameters, enriched with causal mechanisms, through a combination of shared and private branches. 
Besides, we introduce global constraints into the equation to effectively mitigate bias within the various missing domains, thereby elevating the accuracy of our causal effect estimation.
Extensive experiments conducted on new semi-synthetic datasets show that our method outperforms state-of-the-art baselines.
\end{abstract}
\section{Introduction}
%
Causal inference entails the reasoning about relationships between events from data, of which the primary objective is to investigate the impact of interventions on events, establishing genuine causal relationships while avoiding spurious correlations\cite{louizos2017causal,shi2019adapting,zhang2021deep,fan2023interpretable}. For instance, assessing the influence of different medications on patient prognosis\cite{athey2016recursive} or examining the effect of socioeconomic factors on youth employment\cite{frumento2012evaluating}.

In practical scenarios, the data is often dispersed across different silos, requiring a federated approach to estimate causal relationships due to privacy constraints~\cite{mcmahan2017communication,li2020federated,luo2022disentangled,lee2022privacy}. However, differences in data feature dimensions and sample sizes across silos can introduce local biases when estimating causal effects, such as variations in medical records across different hospitals for the same patient. Our aim is to develop a cross-silo causal inference model that works with local data, adapting to variations in feature space and sample size while addressing privacy concerns to some extent.

\noindent\textbf{Related works} Recent years have seen the emergence of machine learning methods for estimating various causal effects\cite{shalit2017estimating,alaa2017bayesian,nie2021quasi,bica2020estimating,zhang2023long}. In single domains, these methods often rely on extensive experimentation and observations with similar spatial distribution of data dimensions \cite{wager2018estimation, kunzel2019metalearners, alaa2018limits}. Inductive approaches such as FlextNet\cite{curth2021inductive} leverage structural similarities among latent outcomes for causal effect estimation. HTCE \cite{bica2022transfer} aids in estimating causal effects in the target domain with assistance from source domain data, but it is limited to specific source and target domains. 
FedCI \cite{vo2022bayesian} and CausalRFF \cite{vo2022adaptive} primarily focus on scenarios where different parties have the same data feature dimensions.
In summary, research on cross-silo causal inference accounting for heterogeneous feature dimensions remains unexplored as of now.


\noindent\textbf{Our method} In light of this, we propose FedDCI that advances cross-silo causal inference by promoting proper causal information sharing. Specifically, we employ the shared branch to extract causal information with consistent dimensions across silos and update their model parameters through server aggregation. Additionally, the specific branch captures client-specific causal information, facilitating the exchange of relevant causal information among different clients through forward and backward propagation, thereby promoting local causal effect estimation. Furthermore, we constrain the model parameters of local shared branch in proximity to the aggregated model parameters. This constraint helps mitigate biases in local causal effects arising from feature heterogeneity. Our contributions are:
\\
$\bullet$ We propose a disentangle framework for joint causal effect estimation, accommodating various causal networks for enhanced flexibility. Besides, we design specific constraints for each disentangle module to reduce the estimation bias. \\
$\bullet$ We propose an optimization strategy to train the disentangle network. Besides, we establish theories showing that our strategy admits asymptotic convergence. \\
$\bullet$ We have extensively evaluated our approach using 
semi-synthetic datasets, where the results show that our method outperforms state-of-the-art baselines.

\section{Methodology}
\subsection{Problem Setting}
We consider there are $K$ parties and each with a local dataset $D^k = {(x^k, y^k, w^k)}$. $x^k=[x^{s,k},x^{p,k}]$ denotes covariates including the shared one $x^{s,k}$ and specific one $x^{p,k}$. $w^k$ is a binary or continuous treatment, and $y^k$ signifies outcomes. Besides, we consider the features are heterogeneous between different parties as $\text{dim}(x^i) \neq \text{dim}(x^j)$ for any two parties $i$ and $j$, where $\text{dim}(\cdot)$ denotes dimensionality. 
%

\subsection{Framework of Design}
In the realm of causal inference, the concept of causal effects is frequently employed to describe the magnitude of outcomes for patients (with covariate x) when comparing the predictions under a no-treatment scenario (T=0) to those under a treatment scenario (T=1). The two distinct outcomes resulting from the presence or absence of treatment are referred to as Potential Outcomes (POs). The other outcome($\mu_1$) of POs can be obtained by adding the size of the causal effect($\tau$) to one of the outcomes($\mu_0$), denoted as $(\mu_1=\mu_0+\tau)$. Consequently, POs share a common underlying structure.
On one hand, this shared structure ensures that individuals under different intervention conditions exhibit similar characteristics before the intervention begins, thus mitigating the impact of selection bias. On the other hand, this shared structure provides a framework for connecting non-iid data observed across multiple domains to underlying causal relationships. Therefore, the shared structure among POs plays a pivotal role in estimating causal effects. Consequently, exploring how to leverage the shared structure among POs for cross-domain causal effect estimation is a critical research question.
To this end, 
we aim to achieve the following objectives:\\
$\bullet$To enhance the causal effect estimation in each target domain using data from shared dimensions across multiple domains, even when the dimensions of covariate X may differ across source domains.\\
$\bullet$To allow each client to keep their data locally and perform model training for causal effect estimation on their own.\\
%
Our objective function is:
\begin{equation}
\label{equation:Global objects}
\min_{\omega}\sum\frac{1}{N}L_k(\omega,\theta,\tau),~
~\tau:=\mathbb{E}[Y(1)-Y(0)|X=x]
\end{equation}
where $L_k$ is the local loss in the $k$-th client and $\tau$ is the expectation of the local causal effect.  $x^k$ is the covariates of the model inputs, and $Y(0)/Y(1)$ denotes the output in the case of treatment $T = 0/1$. $\omega$ and $\theta$ denote the model parameters of causal inference.

        \begin{figure}[t] 
        \begin{center}
        \includegraphics[width=0.42\textwidth]{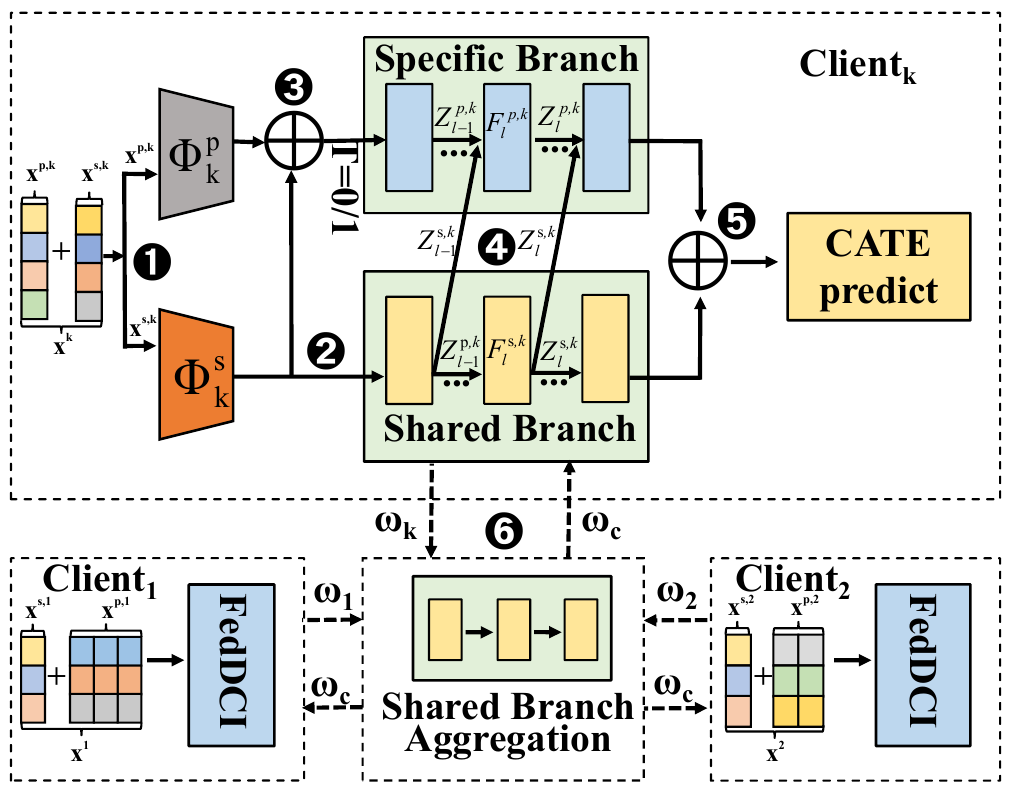}
        \vspace{-8pt}
        \caption{Illustration of our framework. In the local training phase, shared information is communicated to the private branch. 
        During the aggregation stage, the shared model is uploaded to the server for model aggregation.
        }
        \vspace{-20pt}
        \label{fig:Overview}
        \end{center}
        \end{figure}  

\subsection{Model Architecture for Cross-silo Causal Inference}

\textbf{Data Encoder}
%
%
For any client, although the covariates $x^k$ have different distributions $p(x^{s,k})/p(x^{p,k})$ across different clients, they share the same task objective y (causal effect inference). Therefore, our goal is to extract intermediate representations of the target task $z^{s,k}=\phi^{s,k}(x^{s,k})$ and $z^{p,k}=\phi^{p,k}(x^{p,k})$ from different clients. Specifically, we decompose the local objective $p(y|x^{p,k},x^{s,k})$ into $p(y|z^{s,k},z^{p,k})$, $p(z^{s,k}|x^{s,k})$ and $p(z^{p,k}|x^{p,k})$. To maximize the information contained in $z^{s,k}/z^{p,k}$, we aim to bring the posterior distributions $p(z^{s,k}|x^{s,k})$ and $p(z^{p,k}|x^{p,k})$ closer to the distribution $p(z^{*})$, where $z^{*}=\phi(x^{*})$ and $x^{*,k}=[x^{s,k},x^{p,k}]$ to include as much dimensional information as possible. The loss function as: 
\vspace{-5pt}
\begin{equation}
\begin{split}
        L_{ek}&=\min_{\theta^{s,k}} \mathbb{E}_{x\sim p(x^{s})}\left[l_{\text{KL}}\left(p(x^{s,k})p(z^{s,k}|x^{s,k};\theta^{s,k}) \Vert p(z^{*,k})\right)\right]\\
        &+\min_{\theta^{p,k}} \mathbb{E}_{x\sim p(x^{p})}\left[l_{\text{KL}}\left(p(x^{p,k})p(z^{p,k}|x^{p,k};\theta^{p,k}) \Vert p(z^{*,k})\right)\right] \nonumber
\end{split}
\end{equation}
where, $l_{\text{KL}}(\cdot)$ is the KL loss function, $\theta^{s,k}$ is shared encoder parameter and $\theta^{p,k}$ is specific encoder parameter. 


\noindent \textbf{Prediction Model}
%
As shown in Figure~\ref{fig:Overview}, the prediction model includes a specific branch that infers causal relationships from local features, and a shared branch processes common features. For the $k$-th client, $F_l(\cdot)$ is a function of layer $l$, and $z^{s,k}_l$ and $z^{p,k}_l$ represent the shared and private features of layer $l$, respectively. The initial layers use all feature space data to compute $z^{p,k}_{1}$, while the shared branch uses shared features for $z^{s,k}_1$. Intermediate specific feature, denoted as $z^{p,k}_l$, combines the outputs of previous layers from both specific and shared branches, given by $z^{p,k}_l=F_l^p[z^{p,k}_{l-1},z^{s,k}_{l-1}]$, and intermediate shared layers $z^{s,k}_l$ depend solely on the previous shared layer. For example, $z^{s,k}_l=F_l^s[z^{s,k}_{l-1}$]. In the final layer, MLPs are employed to model dimension-1 POs ($\mu_0$ and $\mu_1$). The loss function is:
\begin{equation}
\label{equation:local loss}
L_{w_k}=T_k l_{\text{MSE}}(y_k,\mu_1)-(1-T_k)l_{\text{MSE}}(y_k,\mu_0)+\Vert w_c-w_k\Vert_2^2 \nonumber 
\end{equation}
accounts for treatment ($w$), observed outcomes ($y_k$), and contextual weights (server weigthts $w_c$ and local weights $w_k$). $l_{\text{MSE}}(\cdot)$ represents MSE Loss for continuous or BCE Loss for binary prediction, $T_k$ denotes the probability of intervention, and $\Vert\omega_c-\omega_k\Vert_2^2$ is a global constraint.
%
\subsection{Optimization Strategy}
We use a global training method (Figure \ref{fig:Overview}) that switches between global and local phases. In the local phase, we process data $D^k$ to create shared $z^{s,k}$ and private $z^{p,k}$ representations \ding{182}. Shared features $z^{s,k}$ are common across domains\ding{183} and support cross-silo conditional average treatment effect (CATE) estimation. They go into the shared branch, and concatenated $z^{k} =[z^{s,k},z^{p,k}$] go into specific branches\ding{184}. As training progresses, information from the shared branch transfers to the specific branches\ding{185}, and their outputs estimate PO-based CATE\ding{186}. After each round, shared structures are extracted from data domains via $\omega_k$ aggregation at the server, with global $\omega_c$ returning to enhance information transfer to the specific branches\ding{187}. 


\section{Convergence Analysis}
In this section, we analyze the convergence of our approach, which includes a shared branch and specific branches. Model weights on the central server are denoted as $\omega_t^c$, and client parameters are $\omega_{t}^k = \{\omega_{t}^{p,k}, \omega_{t}^{s,k}\}$, with $\omega_{t}^{p,k}$ for private branch and $\omega_{t}^{s,k}$ for shared branch. 
%
Our convergence analysis is based on the following assumptions.
\begin{asmp}
Non-convexity and L-Lipschitz Smoothness of Objective Function $L_{\omega_k}$:
    \begin{align}\label{eq:theorem1}
         &\Vert L_{\omega_k}(\omega^k_{t+1})-L_{\omega_k}(\omega^k_t)-\langle\nabla L_{\omega_k}(\omega^k_{t}),\omega^k_{t+1}-\omega^k_t\rangle\Vert\nonumber\\
        &\leq \frac{\beta}{2}\Vert\omega^k_t-\omega_{t+1}^k\Vert^2 
    \end{align}
\end{asmp}
\begin{asmp}
    Polyak-Łojasiewicz Property of $\omega_t^{s,k}$ or $\omega_t^{p,k}$:\\ 
         \begin{equation}
           \label{equation:Polyak-Łojasiewicz}
\Vert\nabla L_{\omega_k}(\omega_t^k)\Vert^2 \geq\mu(L_{\omega_k}(\omega_t^k)-L_{\omega_k}(\omega^{k,*}))
        \end{equation}
Additionally, when the local loss functions $L_{\omega_k}$ satisfy the Polyak-Łojasiewicz condition with a positive parameter $l$, it implies that $L_{\omega_k}(\omega)-L_{\omega_k}(\omega^*)\leq \frac{1}{2l}\Vert\nabla L_{\omega_k}(\omega)\Vert^2$, where $\omega^*$ denotes the optimal solution.
\end{asmp}

\begin{theorem}
\label{Theorem:Gradient}
Assuming the validity of assumptions 1, and given that $\Vert\nabla L_{\omega_k}(\omega^{s,k}_t)\Vert^2 \leq A^2$, $\Vert\nabla L_{\omega_k}(\omega^{p,k}_t)\Vert^2 \leq B^2$, and $\xi = \sqrt{\frac{2M}{\beta T(A+B)^2}}$, where $L_{\omega_k}(\omega_1) - L_{\omega_k}(\omega_T) \leq M$, we can demonstrate the following convergence:
         \begin{equation}
           \label{equation:Gradient}
            \min_t \mathbb E_{t\sim T}{\Vert\nabla L_{\omega_k}(\omega_t^{s,k})\Vert^2\leq 2(A+B)\sqrt{\frac{M\beta}{2T}}}
        \end{equation}
Under these conditions, if both $\omega_t^{s,k}$ and $\omega^{p,k}_t$ are smooth, the process can achieve proximity to critical points when the complexity is $O(\frac{1}{\sqrt{T}})$. 
\end{theorem}

\begin{theorem}
           \label{equation:Convergence}
    Moreover, when the Polyak-Łojasiewicz condition is satisfied, we obtain the following convergence bound:
         \begin{equation}
                    \label{equation:Convergence}
         L_{\omega_k}(\omega^{s,k}_{T+1})-L_{\omega_k}(\omega^{s,*})\leq {\frac{2(A+B)}{\mu}\sqrt{\frac{M\beta}{2T}}}
            \end{equation}
Where, $\omega^{s,*}$ represents the optimal model parameters.
\end{theorem}
\section{Experiments}
\subsection{Semi-synthetic Dataset}

To evaluate causal effects, we employ semi-synthetic datasets due to the inherent limitation of not being able to simultaneously observe both counterfactuals and true causal effects for covariates. While existing literature has established benchmarks for domain-specific CATE \cite{ruder2019latent, hill2011bayesian}, there is no standardized benchmark for heterogeneous CATE across multiple domains. To address this gap, we extend the framework of heterogeneous transfer learning introduced by Bica \cite{bica2022transfer} to cross-silo data heterogeneity, allowing us to establish latent connections among distinct domain data. 
Let $x^k$ represent patient features from the $k$-th client dataset, where $d^{s,k}$ and $d^{p,k}$ denote the features of all dimensions in $x^{s,k}$ and $X^{p,k}$ respectively. We propose a concise method for constructing a multi-domain semi-synthetic dataset:
\vspace{-8pt}
\begin{align}\label{equation:Semi-synthetic dataset}
        &Y_k= \epsilon_k+[\frac{\alpha}{K}\sum\limits_{j=1}^{K}\sum\limits_{i=1}^{d^{s,j}} (\omega^{s,j}_i x^{s,j}_i)/d^{s,j}+\\
        &(1-\alpha)[\sum\limits_{i=1}^{d^{p,k}}\beta (\omega^{p,k}_i x_{i}^{p,k})/d^{p,k}+ (1-\beta)\sum\limits_{i=1}^{d^k}(\omega^{k}_ix_i^{k})/d^{k}]] \nonumber
\end{align}
The output $Y_k$ of the POs for the $k$-th client relies on both the specific data $x^k_i$ and the shared data $x^{s,j}$ 
across all domains. $\alpha$ controls the shared structural information proportion between domains in terms of Potential Outcomes (POs), while $\beta$ regulates within-domain shared structural information in terms of POs. Stochasticity is introduced by setting $\omega^{s,k}_{i},\omega^{p,k}_{k},\omega_{i}^k \sim \mathcal{N}(-10, 10)$ and $\epsilon_k \sim \mathcal{N}(0,0.01)$. For each client, considering different $X^k$ values correspond to different treatments, and we allocate treatments using a Bernoulli distribution: $P(W|X) \sim Bernoulli(\gamma(Y(1)-Y(0)))$, where $\gamma$ is the Sigmoid function.

\subsection{Benchmarks Comparison}
\noindent\textbf{Datasets}
The Twins dataset has 11,400 twin pairs with 39 variables, commonly used for causal inference. The IHDP dataset assesses interventions for preterm infants with 747 samples and 25 covariates.\\
\noindent\textbf{Metric}
CATE measures treatment impact on individuals in causal inference, while PEHE represents its error.\\
We conducted a comprehensive study on PEHE within the FedDCI framework. Using both "twins" and "IHDP" datasets, we explored Non-IID data scenarios, evaluating PEHE and ATE. We compared FedDCI with benchmark methods, ensuring fairness by setting $\alpha$ and $\beta$ to 0.5.\\
\textbf{Result 1: Experimental performance of twins for non-independently and identically distributed data}\\
\begin{table}[t]
\centering
 \caption{Results on the twins dataset}
 \label{table:noniid}
 \scalebox{0.7}[0.65]{
\begin{tabular}{@{}lcccccc@{}}
\toprule
\multicolumn{1}{c}{\multirow{2}{*}{Method}} & \multicolumn{3}{c}{The error of CATE}                                                                                                                                                               & \multicolumn{3}{c}{The error of ATE}                                                                                                                                                                \\ \cmidrule(r){2-4} \cmidrule(r){5-7} 
\multicolumn{1}{c}{}                        & \begin{tabular}[c]{@{}c@{}}5\\ clients\end{tabular}             & \begin{tabular}[c]{@{}c@{}}10\\ clients\end{tabular}            & \begin{tabular}[c]{@{}c@{}}15\\ clients\end{tabular}            & \begin{tabular}[c]{@{}c@{}}5\\ clients\end{tabular}             & \begin{tabular}[c]{@{}c@{}}10\\ clients\end{tabular}            & \begin{tabular}[c]{@{}c@{}}15\\ clients\end{tabular}            \\ \cmidrule(r){1-1}\cmidrule(r){2-4} \cmidrule(r){5-7} 
TarNet                                      & \begin{tabular}[c]{@{}c@{}}0.46\\ (±0.02)\end{tabular}          & \begin{tabular}[c]{@{}c@{}}0.25\\ (±0.01)\end{tabular}          & \begin{tabular}[c]{@{}c@{}}0.37\\ (±0.01)\end{tabular}          & \begin{tabular}[c]{@{}c@{}}0.30\\ (±0.01)\end{tabular}          & \begin{tabular}[c]{@{}c@{}}0.05\\ (±0.02)\end{tabular}          & \begin{tabular}[c]{@{}c@{}}0.21\\ (±0.02)\end{tabular}          \\
TNet                                        & \begin{tabular}[c]{@{}c@{}}0.66\\ (±0.01)\end{tabular}          & \begin{tabular}[c]{@{}c@{}}0.26\\ (±0.01)\end{tabular}          & \begin{tabular}[c]{@{}c@{}}0.49\\ (±0.04)\end{tabular}          & \begin{tabular}[c]{@{}c@{}}0.47\\ (±0.02)\end{tabular}          & \begin{tabular}[c]{@{}c@{}}0.09\\ (±0.01)\end{tabular}          & \begin{tabular}[c]{@{}c@{}}0.34\\ (±0.01)\end{tabular}          \\
SNet                                        & \begin{tabular}[c]{@{}c@{}}0.45\\ (±0.03)\end{tabular}          & \begin{tabular}[c]{@{}c@{}}0.26\\ (±0.01)\end{tabular}          & \begin{tabular}[c]{@{}c@{}}0.53\\ (±0.01)\end{tabular}          & \begin{tabular}[c]{@{}c@{}}0.18\\ (±0.01)\end{tabular}          & \begin{tabular}[c]{@{}c@{}}0.09\\ (±0.01)\end{tabular}          & \begin{tabular}[c]{@{}c@{}}0.38\\ (±0.01)\end{tabular}          \\
DRLearner                                   & \begin{tabular}[c]{@{}c@{}}0.41\\ (±0.02)\end{tabular}          & \begin{tabular}[c]{@{}c@{}}0.27\\ (±0.02\end{tabular}           & \begin{tabular}[c]{@{}c@{}}0.57\\ (±0.01)\end{tabular}          & \begin{tabular}[c]{@{}c@{}}0.18\\ (±0.02)\end{tabular}          & \begin{tabular}[c]{@{}c@{}}0.11\\ (±0.01)\end{tabular}          & \begin{tabular}[c]{@{}c@{}}0.45\\ (±0.01)\end{tabular}          \\
PWLearner                                   & \begin{tabular}[c]{@{}c@{}}0.39\\ (±0.01)\end{tabular}          & \begin{tabular}[c]{@{}c@{}}0.27\\ (±0.02)\end{tabular}          & \begin{tabular}[c]{@{}c@{}}0.59\\ (±0.01)\end{tabular}          & \begin{tabular}[c]{@{}c@{}}0.16\\ (±0.01)\end{tabular}          & \begin{tabular}[c]{@{}c@{}}0.10\\ (±0.01)\end{tabular}          & \begin{tabular}[c]{@{}c@{}}0.47\\ (±0.00)\end{tabular}          \\
RALearner                                   & \begin{tabular}[c]{@{}c@{}}0.45\\ (±0.01)\end{tabular}          & \begin{tabular}[c]{@{}c@{}}0.27\\ (±0.03)\end{tabular}          & \begin{tabular}[c]{@{}c@{}}0.63\\ (±0.01)\end{tabular}          & \begin{tabular}[c]{@{}c@{}}0.25\\ (±0.01)\end{tabular}          & \begin{tabular}[c]{@{}c@{}}0.11\\ (±0.01)\end{tabular}          & \begin{tabular}[c]{@{}c@{}}0.52\\ (±0.01)\end{tabular}          \\
CausalRFF                                   & \begin{tabular}[c]{@{}c@{}}0.53\\ (±0.01)\end{tabular}          & \begin{tabular}[c]{@{}c@{}}0.61\\ (±0.03)\end{tabular}          & \begin{tabular}[c]{@{}c@{}}1.18\\ (±0.01)\end{tabular}          & \begin{tabular}[c]{@{}c@{}}0.15\\ (±0.00)\end{tabular}          & \begin{tabular}[c]{@{}c@{}}0.47\\ (±0.01)\end{tabular}          & \begin{tabular}[c]{@{}c@{}}0.13\\ (±0.01)\end{tabular}          \\
FedCI                                    & \begin{tabular}[c]{@{}c@{}}0.46\\ (±0.03)\end{tabular}          & \begin{tabular}[c]{@{}c@{}}0.57\\ (±0.03)\end{tabular}          & \begin{tabular}[c]{@{}c@{}}1.05\\ (±0.01)\end{tabular}          & \begin{tabular}[c]{@{}c@{}}0.08\\ (±0.02)\end{tabular}          & \begin{tabular}[c]{@{}c@{}}0.12\\ (±0.02)\end{tabular}          & \begin{tabular}[c]{@{}c@{}}0.18\\ (±0.01)\end{tabular}          \\\cmidrule(r){1-1}\cmidrule(r){2-4} \cmidrule(r){5-7} 
FedDCI                                      & \textbf{\begin{tabular}[c]{@{}c@{}}0.34\\ (±0.01)\end{tabular}} & \textbf{\begin{tabular}[c]{@{}c@{}}0.25\\ (±0.01)\end{tabular}} & \textbf{\begin{tabular}[c]{@{}c@{}}0.29\\ (±0.01)\end{tabular}} & \textbf{\begin{tabular}[c]{@{}c@{}}0.12\\ (±0.01)\end{tabular}} & \textbf{\begin{tabular}[c]{@{}c@{}}0.04\\ (±0.01)\end{tabular}} & \textbf{\begin{tabular}[c]{@{}c@{}}0.11\\ (±0.02)\end{tabular}} \\ \bottomrule
\end{tabular}
}
\vspace{-18pt}
\end{table}
\begin{figure}[b!]
    \vspace{-20pt}
    \centering
    \subfigure[The PEHE performance of different methods.]{
        \centering
        \begin{minipage}[t]{1\columnwidth}
            \includegraphics[width=0.8\columnwidth]{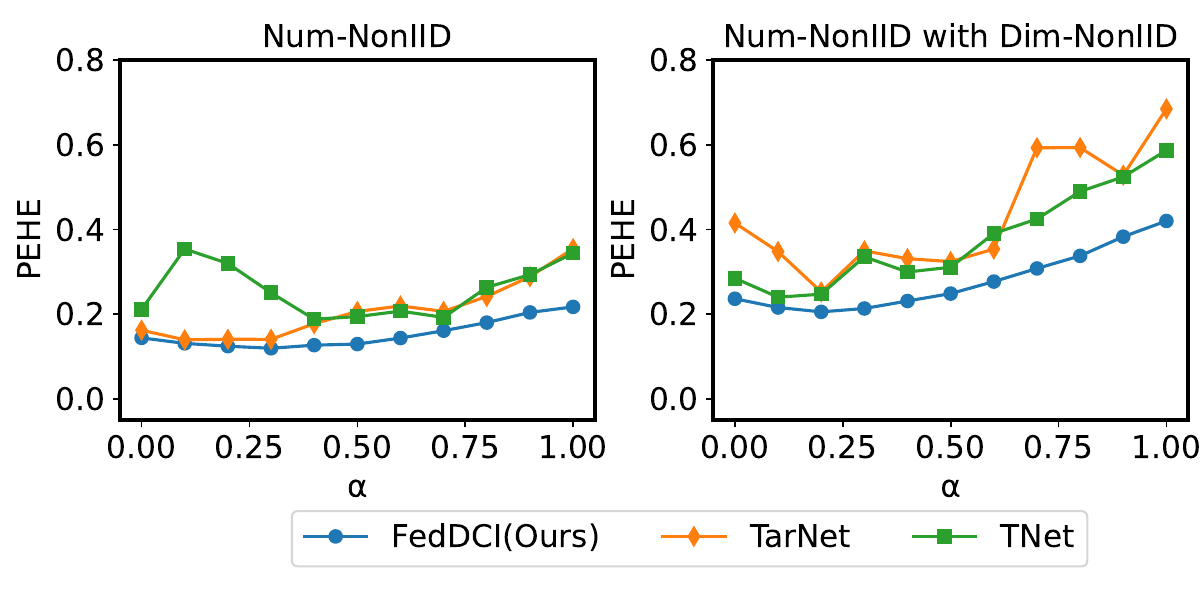}
        \end{minipage}%

    }
    \subfigure[The ATE performance of different methods.]{
        \centering
        \begin{minipage}[t]{1\columnwidth}
            \includegraphics[width=0.8\columnwidth]{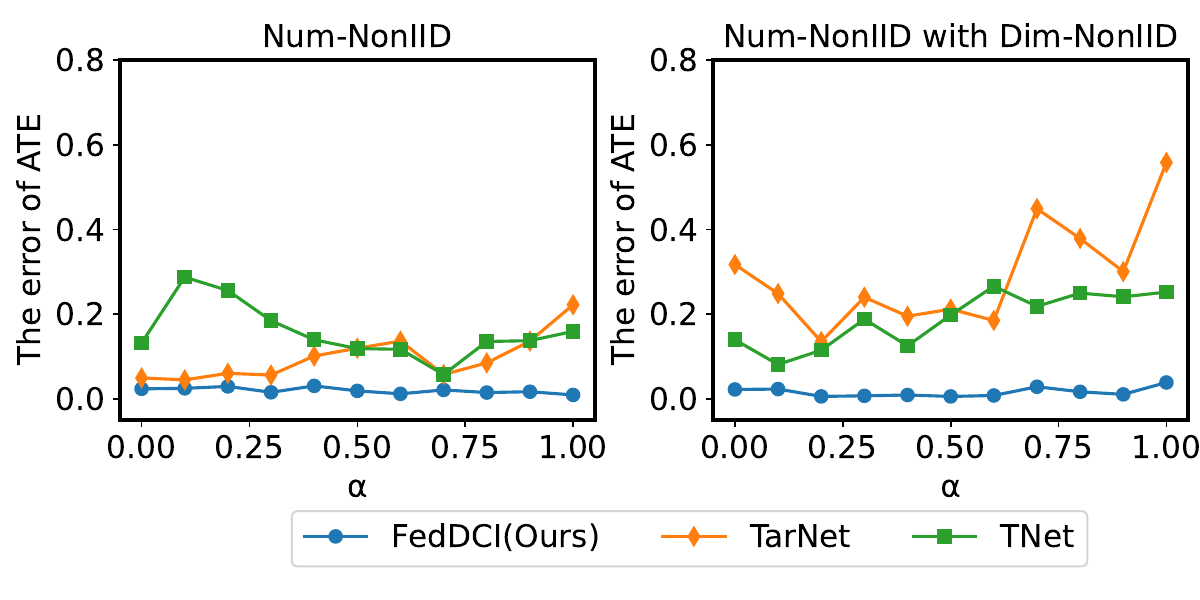}
        \end{minipage}%
    }
    \centering
       \vspace{-16pt}
    \caption{This experiment analysed the effect of $\alpha$ on PEHE and ATE metrics in the case of sample size noniid versus in the case of sample size, sample characteristics both non-iid.}
    \label{fig:alpha_influence}
    \vskip -0.3in
\end{figure}
\noindent FedDCI shines in handling Non-Identically Distributed (Non-IID) data, as seen in Table \ref{table:noniid}. It excels in extracting valuable insights from diverse client data, combining shared and unique attributes. Through astute aggregation of model features, it effectively uncovers individual client-specific information patterns, leading to a more refined understanding of causal relationships in this complex context.\\
\noindent\textbf{Result 2: Experimental performance of IHDP for non-independently and identically distributed data}\\
We present a summary of the experimental outcomes conducted on the semi-synthetic dataset IHDP using the formulation referenced as Eq. \ref{equation:Semi-synthetic dataset}. The results are presented in Table \ref{table:IHDPnoniid}. Our experimentation encompassed two key aspects. Firstly, we conducted training within a federated framework, which allowed us to retain the dataset locally and minimize the risks associated with privacy breaches. Secondly, we extended our evaluation to a spatially heterogeneous federated setting, focusing on cross-silo causal effects.

\begin{table}[t]
\centering
 \caption{Results on the IHDP dataset}
 \label{table:IHDPnoniid}
  \scalebox{0.7}[0.65]{
\begin{tabular}{@{}lcccccc@{}}
\toprule
\multirow{2}{*}{Method} & \multicolumn{3}{c}{The error of CATE}                                                                                                                                                             & \multicolumn{3}{c}{The error of ATE}                                                                                                                                                             \\ \cmidrule(r){2-4} \cmidrule(r){5-7}
                        & \begin{tabular}[c]{@{}c@{}}5\\ clients\end{tabular}            & \begin{tabular}[c]{@{}c@{}}10\\ clients\end{tabular}            & \begin{tabular}[c]{@{}c@{}}15\\ clients\end{tabular}           & \begin{tabular}[c]{@{}c@{}}5\\ clients\end{tabular}            & \begin{tabular}[c]{@{}c@{}}10\\ clients\end{tabular}           & \begin{tabular}[c]{@{}c@{}}15\\ clients\end{tabular}           \\ \cmidrule(r){1-1}\cmidrule(r){2-4} \cmidrule(r){5-7}
TarNet                  & \begin{tabular}[c]{@{}c@{}}1.00\\ (±0.02)\end{tabular}          & \begin{tabular}[c]{@{}c@{}}1.35\\ (±0.08)\end{tabular}           & \begin{tabular}[c]{@{}c@{}}1.58\\ (±0.09)\end{tabular}          & \begin{tabular}[c]{@{}c@{}}0.54\\ (±0.03)\end{tabular} & \begin{tabular}[c]{@{}c@{}}1.07\\ (±0.02)\end{tabular}          & \begin{tabular}[c]{@{}c@{}}1.23\\ (±0.09)\end{tabular}          \\
TNet                    & \begin{tabular}[c]{@{}c@{}}1.12\\ (±0.08)\end{tabular}          & \begin{tabular}[c]{@{}c@{}}1.90\\ (±0.01)\end{tabular}            & \begin{tabular}[c]{@{}c@{}}1.57\\ (±0.08\end{tabular}           & \begin{tabular}[c]{@{}c@{}}0.64\\ (±0.06)\end{tabular}          & \begin{tabular}[c]{@{}c@{}}1.32\\ (±0.07)\end{tabular}          & \begin{tabular}[c]{@{}c@{}}1.12\\ (±0.01)\end{tabular}          \\
SNet                    & \begin{tabular}[c]{@{}c@{}}1.28\\ (±0.07)\end{tabular}          & \begin{tabular}[c]{@{}c@{}}1.59\\ (±0.02)\end{tabular}           & \begin{tabular}[c]{@{}c@{}}1.19\\ (±0.03)\end{tabular}          & \begin{tabular}[c]{@{}c@{}}0.77\\ (±0.04)\end{tabular}          & \begin{tabular}[c]{@{}c@{}}1.07\\ (±0.01)\end{tabular}          & \begin{tabular}[c]{@{}c@{}}0.71\\ (±0.05)\end{tabular}          \\
DRLearner               & \begin{tabular}[c]{@{}c@{}}1.28\\ (±0.02)\end{tabular}          & \begin{tabular}[c]{@{}c@{}}1.19\\ (±0.05)\end{tabular}           & \begin{tabular}[c]{@{}c@{}}1.17\\ (±0.05)\end{tabular}          & \begin{tabular}[c]{@{}c@{}}0.94\\ (±0.08)\end{tabular}          & \begin{tabular}[c]{@{}c@{}}0.92\\ (±0.04)\end{tabular}          & \textbf{\begin{tabular}[c]{@{}c@{}}0.57\\ (0.08)\end{tabular}} \\
PWLearner               & \begin{tabular}[c]{@{}c@{}}1.00\\ (±0.01)\end{tabular}          & \begin{tabular}[c]{@{}c@{}}1.34\\ (±0.01)\end{tabular}           & \begin{tabular}[c]{@{}c@{}}1.29\\ (±0.01)\end{tabular}          & \begin{tabular}[c]{@{}c@{}}0.72\\ (±0.05)\end{tabular}          & \begin{tabular}[c]{@{}c@{}}1.04\\ (±0.08)\end{tabular}          & \begin{tabular}[c]{@{}c@{}}0.66\\ (±0.01)\end{tabular}          \\
RALearner               & \begin{tabular}[c]{@{}c@{}}1.08\\ (±0.04)\end{tabular}          & \begin{tabular}[c]{@{}c@{}}1.22\\ (±0.09)\end{tabular}           & \begin{tabular}[c]{@{}c@{}}1.20\\ (±0.04)\end{tabular}          & \begin{tabular}[c]{@{}c@{}}0.66\\ (±0.06)\end{tabular}          & \begin{tabular}[c]{@{}c@{}}0.99\\ (±0.09)\end{tabular}          & \begin{tabular}[c]{@{}c@{}}0.75\\ (±0.05)\end{tabular}          \\
CausalRFF               & \begin{tabular}[c]{@{}c@{}}1.39\\ (±0.09)\end{tabular}          & \begin{tabular}[c]{@{}c@{}}1.29\\ (±0.04)\end{tabular}           & \begin{tabular}[c]{@{}c@{}}1.52\\ (±0.07)\end{tabular}          & \begin{tabular}[c]{@{}c@{}}0.95\\ (±0.05)\end{tabular}          & \begin{tabular}[c]{@{}c@{}}0.87\\ (±0.09)\end{tabular}          & \begin{tabular}[c]{@{}c@{}}0.99\\ (±0.08)\end{tabular}          \\ 
FedCI               & \begin{tabular}[c]{@{}c@{}}1.42\\ (±0.03)\end{tabular}          & \textbf{\begin{tabular}[c]{@{}c@{}}1.13\\ (±0.06)\end{tabular}}           & \begin{tabular}[c]{@{}c@{}}1.17\\ (±0.03)\end{tabular}          & \textbf{\begin{tabular}[c]{@{}c@{}}0.41\\ (±0.02)\end{tabular}}          & \begin{tabular}[c]{@{}c@{}}0.82\\ (±0.03)\end{tabular}          & \begin{tabular}[c]{@{}c@{}}0.70\\ (±0.01)\end{tabular}          \\\cmidrule(r){1-1}\cmidrule(r){2-4} \cmidrule(r){5-7}
FedDCI                  & \textbf{\begin{tabular}[c]{@{}c@{}}0.96\\ (±0.05)\end{tabular}} & \begin{tabular}[c]{@{}c@{}}1.18\\ (±0.03)\end{tabular} & \textbf{\begin{tabular}[c]{@{}c@{}}1.04\\ (±0.01)\end{tabular}} & \begin{tabular}[c]{@{}c@{}}0.55\\ (±0.01)\end{tabular}          & \textbf{\begin{tabular}[c]{@{}c@{}}0.75\\ (±0.01)\end{tabular}} & \begin{tabular}[c]{@{}c@{}}0.61\\ (±0.02)\end{tabular}          \\ \bottomrule
\end{tabular}
}
\vspace{-13pt}
\end{table}
\vspace{-8pt}
\subsection{Impact of Shared Ratio on Potential Outcomes}
FedDCI outperforms baseline methods on the IHDP dataset (Table \ref{table:IHDPnoniid}), excelling in predicting CATE with lower error rates across clients. When considering ATE, FedDCI maintained consistent errors with the baseline, confirming its reliability.

For diverse client datasets (non-iid features), FedDCI adapted well. Figure \ref{fig:alpha_influence} demonstrates that in IID data, varying $\alpha$ affected the single-branch network TNet more than in Non-IID data. FedDCI showed superior adaptability to different $\alpha$ values with lower PEHE scores and smoother curves. Importantly, it consistently achieved lower ATE errors across various $\alpha$ scenarios, indicating its precision in capturing cross-silo causal relationships.
\vspace{-3pt}
\section{Conclusion}
We propose a method for estimating causal effects across diverse domains in a heterogeneous space. Our approach enhances causal effect estimation in the target domain by leveraging inter-domain correlations from distinct feature spaces while maintaining data locality. We introduce an improved flexible disentangle framework that transfers model parameters across domains through shared and private branches, enabling us to estimate causal effects across diverse domains. We conduct extensive experiments over different datasets and demonstrate the effectiveness of the proposed method.
\vspace{-3pt}
\ninept
\subsection*{Acknowledgments}
This work is supported by National Natural Science Foundation of China under grants 62302184. The author would like to thank Nuowei Technology for their support.


\bibliographystyle{IEEEbib}
\bibliography{strings,refs}

\begin{thebibliography}{10}

\bibitem{louizos2017causal}
Christos Louizos, Uri Shalit, Joris~M Mooij, David Sontag, Richard Zemel, and
  Max Welling,
\newblock ``Causal effect inference with deep latent-variable models,''
\newblock {\em Advances in neural information processing systems}, vol. 30,
  2017.

\bibitem{shi2019adapting}
Claudia Shi, David Blei, and Victor Veitch,
\newblock ``Adapting neural networks for the estimation of treatment effects,''
\newblock {\em Advances in neural information processing systems}, vol. 32,
  2019.

\bibitem{zhang2021deep}
Xingxuan Zhang, Peng Cui, Renzhe Xu, Linjun Zhou, Yue He, and Zheyan Shen,
\newblock ``Deep stable learning for out-of-distribution generalization,''
\newblock in {\em Proceedings of the IEEE/CVF Conference on Computer Vision and
  Pattern Recognition}, 2021, pp. 5372--5382.

\bibitem{fan2023interpretable}
Chenchen Fan, Yixin Wang, Yahong Zhang, and Wenli Ouyang,
\newblock ``Interpretable multi-scale neural network for granger causality
  discovery,''
\newblock in {\em ICASSP 2023-2023 IEEE International Conference on Acoustics,
  Speech and Signal Processing (ICASSP)}. IEEE, 2023, pp. 1--5.

\bibitem{athey2016recursive}
Susan Athey and Guido Imbens,
\newblock ``Recursive partitioning for heterogeneous causal effects,''
\newblock {\em Proceedings of the National Academy of Sciences}, vol. 113, no.
  27, pp. 7353--7360, 2016.

\bibitem{frumento2012evaluating}
Paolo Frumento, Fabrizia Mealli, Barbara Pacini, and Donald~B Rubin,
\newblock ``Evaluating the effect of training on wages in the presence of
  noncompliance, nonemployment, and missing outcome data,''
\newblock {\em Journal of the American Statistical Association}, vol. 107, no.
  498, pp. 450--466, 2012.

\bibitem{mcmahan2017communication}
Brendan McMahan, Eider Moore, Daniel Ramage, Seth Hampson, and Blaise~Aguera
  y~Arcas,
\newblock ``Communication-efficient learning of deep networks from
  decentralized data,''
\newblock in {\em Artificial intelligence and statistics}. PMLR, 2017, pp.
  1273--1282.

\bibitem{li2020federated}
Tian Li, Anit~Kumar Sahu, Manzil Zaheer, Maziar Sanjabi, Ameet Talwalkar, and
  Virginia Smith,
\newblock ``Federated optimization in heterogeneous networks,''
\newblock {\em Proceedings of Machine learning and systems}, vol. 2, pp.
  429--450, 2020.

\bibitem{luo2022disentangled}
Zhengquan Luo, Yunlong Wang, Zilei Wang, Zhenan Sun, and Tieniu Tan,
\newblock ``Disentangled federated learning for tackling attributes skew via
  invariant aggregation and diversity transferring,''
\newblock in {\em International Conference on Machine Learning}. PMLR, 2022,
  pp. 14527--14541.

\bibitem{lee2022privacy}
Harlin Lee, Andrea~L Bertozzi, Jelena Kova{\v{c}}evi{\'c}, and Yuejie Chi,
\newblock ``Privacy-preserving federated multi-task linear regression: A
  one-shot linear mixing approach inspired by graph regularization,''
\newblock in {\em ICASSP 2022-2022 IEEE International Conference on Acoustics,
  Speech and Signal Processing (ICASSP)}. IEEE, 2022, pp. 5947--5951.

\bibitem{shalit2017estimating}
Uri Shalit, Fredrik~D Johansson, and David Sontag,
\newblock ``Estimating individual treatment effect: generalization bounds and
  algorithms,''
\newblock in {\em International conference on machine learning}. PMLR, 2017,
  pp. 3076--3085.

\bibitem{alaa2017bayesian}
Ahmed~M Alaa and Mihaela Van Der~Schaar,
\newblock ``Bayesian inference of individualized treatment effects using
  multi-task gaussian processes,''
\newblock {\em Advances in neural information processing systems}, vol. 30,
  2017.

\bibitem{nie2021quasi}
Xinkun Nie and Stefan Wager,
\newblock ``Quasi-oracle estimation of heterogeneous treatment effects,''
\newblock {\em Biometrika}, vol. 108, no. 2, pp. 299--319, 2021.

\bibitem{bica2020estimating}
Ioana Bica, James Jordon, and Mihaela van~der Schaar,
\newblock ``Estimating the effects of continuous-valued interventions using
  generative adversarial networks,''
\newblock {\em Advances in Neural Information Processing Systems}, vol. 33, pp.
  16434--16445, 2020.

\bibitem{zhang2023long}
Yahong Zhang, Sheng Shi, ChenChen Fan, Yixin Wang, Wenli Ouyang, Jianpin Fan,
  et~al.,
\newblock ``Long-tailed recognition with causal invariant transformation,''
\newblock in {\em ICASSP 2023-2023 IEEE International Conference on Acoustics,
  Speech and Signal Processing (ICASSP)}. IEEE, 2023, pp. 1--5.

\bibitem{wager2018estimation}
Stefan Wager and Susan Athey,
\newblock ``Estimation and inference of heterogeneous treatment effects using
  random forests,''
\newblock {\em Journal of the American Statistical Association}, vol. 113, no.
  523, pp. 1228--1242, 2018.

\bibitem{kunzel2019metalearners}
S{\"o}ren~R K{\"u}nzel, Jasjeet~S Sekhon, Peter~J Bickel, and Bin Yu,
\newblock ``Metalearners for estimating heterogeneous treatment effects using
  machine learning,''
\newblock {\em Proceedings of the national academy of sciences}, vol. 116, no.
  10, pp. 4156--4165, 2019.

\bibitem{alaa2018limits}
Ahmed Alaa and Mihaela Schaar,
\newblock ``Limits of estimating heterogeneous treatment effects: Guidelines
  for practical algorithm design,''
\newblock in {\em International Conference on Machine Learning}. PMLR, 2018,
  pp. 129--138.

\bibitem{curth2021inductive}
Alicia Curth and Mihaela van~der Schaar,
\newblock ``On inductive biases for heterogeneous treatment effect
  estimation,''
\newblock {\em Advances in Neural Information Processing Systems}, vol. 34, pp.
  15883--15894, 2021.

\bibitem{bica2022transfer}
Ioana Bica and Mihaela van~der Schaar,
\newblock ``Transfer learning on heterogeneous feature spaces for treatment
  effects estimation,''
\newblock {\em Advances in Neural Information Processing Systems}, vol. 35, pp.
  37184--37198, 2022.

\bibitem{vo2022bayesian}
Thanh~Vinh Vo, Young Lee, Trong~Nghia Hoang, and Tze-Yun Leong,
\newblock ``Bayesian federated estimation of causal effects from observational
  data,''
\newblock in {\em Uncertainty in Artificial Intelligence}. PMLR, 2022, pp.
  2024--2034.

\bibitem{vo2022adaptive}
Thanh~Vinh Vo, Arnab Bhattacharyya, Young Lee, and Tze-Yun Leong,
\newblock ``An adaptive kernel approach to federated learning of heterogeneous
  causal effects,''
\newblock {\em Advances in Neural Information Processing Systems}, vol. 35, pp.
  24459--24473, 2022.

\bibitem{ruder2019latent}
Sebastian Ruder, Joachim Bingel, Isabelle Augenstein, and Anders S{\o}gaard,
\newblock ``Latent multi-task architecture learning,''
\newblock in {\em Proceedings of the AAAI Conference on Artificial
  Intelligence}, 2019, vol.~33, pp. 4822--4829.

\bibitem{hill2011bayesian}
Jennifer~L Hill,
\newblock ``Bayesian nonparametric modeling for causal inference,''
\newblock {\em Journal of Computational and Graphical Statistics}, vol. 20, no.
  1, pp. 217--240, 2011.

\end{thebibliography}

\newpage
\subsection*{APPENDIX}
\noindent\textbf{Proof of Convergence Analysis}\\
In this section, we present the convergence analysis of our proposed optimization process, which is composed of a shared branch and a specific branch, compared to the traditional FedAvg. We define $\omega_t^c$ as the model weights on the central server, and the model parameters on each client are denoted as $\omega^k_{t} = \{\omega_{t}^{p,k}, \omega_{t}^{s,k}\}$, where $\omega_{t}^{p,k}$ represents the parameters of the private branch and $\omega_{t}^{s,k}$ represents the parameters of the shared branch. After each round of server aggregation, the shared branch $\omega_{t}^{s,k}$ is updated from $\omega_t^c$. We base our convergence analysis on the following assumptions.
The model parameters are assumed to be $\omega_t^k=\{\omega_t^{s,k},\omega_t^{p,k}\}.$\\
Let $x=\omega_{t+1}^k,y=\omega_{t}^k$ and the gradient update be:
\setcounter{equation}{0}
\begin{align}
\label{proof:gradient}
\omega^k_{t+1}=\omega_t^k-\eta \nabla L_{\omega_k}(\omega_t^k)
\end{align}
and $\nabla L_{\omega_k}(\omega^k_t)=\nabla L_{\omega_k}(\omega^{s,k}_t)+L_{\omega_k}(\omega^{p,k}_t)$\\
According to the smooth assumption, Eq. \ref{proof:gradient} is obtained by substituting Eq:
\begin{align}
\label{proof:smooth}
&L_{\omega_k}(\omega_{t+1}^k)-L_{\omega_k}(\omega_t^k)+\langle\nabla L_{\omega_k}(\omega_t^k),\omega_{t+1}^k-\omega_t^k\rangle\nonumber\\
&\leq \frac{\beta}{2}\Vert\omega_{t}^k-\omega_{t+1}^k\Vert^2
\end{align}
If $\Vert L_{\omega_k}(\omega_t^{s,k})\Vert^2\leq A$ and $\Vert L_{\omega_k}(\omega_t^{p,k})\Vert^2\leq B$ then $\Vert L_{\omega_k}(\omega_t^k)\Vert^2\leq A+B$, we have:
\begin{align}
    &L_{\omega_k}(\omega_{t+1}^k)-L_{\omega_k}(\omega_t^k)\nonumber\\
    &+\eta\langle\nabla L_{\omega_k}(\omega_t^k),(\nabla L_{\omega_k}(\omega_t^{s,k})+\nabla L_{\omega_k}(\omega_t^{p,k}))\rangle  \leq \frac{\beta \eta^2}{2}(A+B)
\end{align}
Then take an expectation on the value:
\begin{align}
  &\mathbb E [L_{\omega_k}(\omega_{t+1}^k)-L_{\omega_k}(\omega_t^k)]+
  \mathbb E [\eta\Vert\nabla L_{\omega_k}(\omega_t^{s,k})+\nabla L_{\omega_k}(\omega_t^{p,k}))\Vert^2]\nonumber\\
  &\leq \frac{\beta \eta^2}{2}(A+B)  
\end{align}
Accumulating from t = 1 to T yields:
\begin{align}
    &\mathbb E [L_{\omega_k}(\omega_{T}^k)-L_{\omega_k}(\omega_1^k)]+\sum_{t=1}^T\mathbb E[\eta\Vert\nabla L_{\omega_k}(\omega_t^{s,k})+\nabla L_{\omega_k}(\omega_t^{p,k}))\Vert^2] \nonumber\\
    &\leq \frac{\beta \eta^2 T}{2}(A+B)
\end{align}
At this point if $\Vert L_{\omega_k}(\omega_T^k)-L_{\omega_k}(\omega_1^k)\Vert\leq M$ can be introduced:
\begin{align}
    \sum_{t=1}^T\Vert\nabla L_{\omega_k}(\omega_t^{s,k})+\nabla L_{\omega_k}(\omega_t^{p,k})\Vert^2\leq \frac{\beta \eta T\Vert A+B\Vert^2}{2}+\frac{M}{\eta}
\end{align}
Also divide by T to obtain:
\begin{align}
    E\Vert\nabla L_{\omega_k}(\omega_t^{s,k})+\nabla L_{\omega_k}(\omega_t^{p,k})\Vert^2\leq \frac{\beta \eta \Vert A+B\Vert^2}{2}+\frac{M}{T\eta}
\end{align}
Since $\nabla L_{\omega_k}(\omega_t^{p,k})\leq B$, therefore:
\begin{align}
    \mathbb E\Vert\nabla L_{\omega_k}(\omega_t^{s,k})\Vert&\leq \sqrt{\frac{\beta \eta \Vert A+B\Vert^2}{2}+\frac{M}{T\eta}}-\mathbb E\Vert \nabla L_{\omega_k}(\omega_t^{p,k})\Vert\nonumber\\
    &\leq \sqrt{\frac{\beta \eta \Vert A+B\Vert^2}{2}+\frac{M}{T\eta}}
\end{align}
When $\eta=\sqrt{2M}{\beta T(A+B)^2}$ is obtained:
\begin{align}
    \label{proof:gradient bound}
    \min_t \mathbb E_{t\sim T}{\Vert\nabla L_{\omega_k}(\omega_t^{s,k})\Vert^2\leq {2(A+B)\sqrt{\frac{M\beta}{2T}}}}
\end{align}
Take Eq.\ref{proof:gradient} in Eq.\ref{proof:gradient bound}, we have:
\begin{align}
   \label{proof:gradient bound}
     L_{\omega_k}(\omega^{s,k}_{T+1})-L_{\omega_k}(\omega^{s,*})\leq {\frac{2(A+B)}{\mu}\sqrt{\frac{M\beta}{2T}}}
\end{align}
\end{document}